\documentclass[letterpaper, 10 pt, journal, twoside]{IEEEtran}  

\IEEEoverridecommandlockouts                              

\usepackage{cite}
\usepackage{amsmath,amssymb,amsfonts}
\usepackage{algorithmic}
\usepackage{graphicx}
\usepackage{textcomp}
\usepackage{xcolor}
\usepackage{bm}
\usepackage{balance}
\usepackage{multirow}
\usepackage{makecell}
\usepackage{booktabs}
\usepackage{threeparttable}
\usepackage{pifont}
\usepackage{soul}
\usepackage[pagewise]{lineno}
\soulregister{\cite}7
\soulregister{\ref}7
\soulregister{\ding}7

\author{Shiwei Lian$^{1}$, and Feitian Zhang$^{2}$
\thanks{This manuscript is accepted by IEEE Robotics and Automation Letters. \it{(Corresponding Author: Feitian Zhang.)}} 
\thanks{$^{1}$Shiwei Lian is with the Department of Advanced Manufacturing and Robotics, College of Engineering, Peking University, Beijing, 100871, China
        {\tt\footnotesize lianshiwei@stu.pku.edu.cn}}%
\thanks{$^{2}$Feitian Zhang is with the Department of Advanced Manufacturing and Robotics, and the State Key Laboratory of Turbulence and Complex Systems, College of Engineering, Peking University, Beijing, 100871, China
        {\tt\footnotesize feitian@pku.edu.cn}}%
}

\title{TDANet: Target-Directed Attention Network For Object-Goal Visual Navigation With Zero-Shot Ability}

\begin{document}

\maketitle

\begin{abstract}
The generalization of the end-to-end deep reinforcement learning (DRL) for object-goal visual navigation is a long-standing challenge since object classes and placements vary in new test environments. Learning domain-independent visual representation is critical for enabling the trained DRL agent with the ability to generalize to unseen scenes and objects. In this letter, a target-directed attention network (TDANet) is proposed to learn the end-to-end object-goal visual navigation policy with zero-shot ability. TDANet features a novel target attention (TA) module that learns both the spatial and semantic relationships among objects to help TDANet focus on the most relevant observed objects to the target. With the Siamese architecture (SA) design, TDANet distinguishes the difference between the current and target states and generates the domain-independent visual representation. To evaluate the navigation performance of TDANet, extensive experiments are conducted in the AI2-THOR embodied AI environment. The simulation results demonstrate a strong generalization ability of TDANet to unseen scenes and target objects, with higher navigation success rate (SR) and success weighted by length (SPL) than other state-of-the-art models. TDANet is finally deployed on a wheeled robot in real scenes, demonstrating satisfactory generalization of TDANet to the real world.
\end{abstract}



\section{Introduction}
\IEEEPARstart{T}{he} objective of object-goal visual navigation is to find a target object in an environment using only egocentric visual observations. This task poses a major challenge to robots, requiring their visual understanding and inference of the complex scene to successfully locate a target instance. 

Recent studies \cite{ORG, Lyu2022ImprovingTV, VTNet, OMT, SP, mjol, Li, ral, SSNet} have achieved great advancement in solving this problem using deep reinforcement learning (DRL) to train an end-to-end model. Learning an informative visual representation containing the relationships among objects is of crucial importance to the design of a robust navigation policy\cite{ORG}. The learning of domain-independent feature encoding relationships among objects is a long-standing ill-posed problem\cite{Lyu2022ImprovingTV} due to the existence of irrelevant information in RGB images such as background textures and colors. In addition, test environments vary in object classes and placements further complicating the deployment of the trained policy. Du \textit{et al.} \cite{VTNet} and Fukushima \textit{et al.} \cite{OMT} applied visual transformer\cite{vt} to exploit the relationships of detected instances, which, however, at the same time increased the learning difficulty and computational cost. Other studies utilized prior knowledge graphs \cite{SP, mjol, Li} or cosine similarity of word embeddings \cite{ral, OMT} to assist in learning navigation strategies. These prior knowledge or representations of object relationships are rarely updated during training, thus limiting the adaptation to unseen scenes with dissimilar object placements. Moreover, most end-to-end models mainly focus on a limited class of target objects\cite{SSNet}, while in the real household environment there may exist target objects not seen during training.

This letter investigates the generalization of the DRL-based object-goal visual navigation to both unseen test environments and unseen target objects by learning the relationships among objects. A target-directed attention network (TDANet) is proposed to train an end-to-end DRL policy for object-goal visual navigation with zero-shot ability. The network focuses on objects in the current visual observation that show strong correspondence with the target. A novel target attention (TA) module uses the information of the observed and target objects as well as their word embeddings as the input to learn both the spatial and semantic relationships between the target object and the detected objects in training. TDANet adopts the design of the Siamese architecture (SA) and distinguishes the difference between the current and desired states of the agent to guide the navigation, demonstrating strong zero-shot ability. The proposed model is evaluated in the AI2-THOR \cite{ai2thor} embodied AI environment. The simulation results show that TDANet generalizes satisfactorily to unseen objects and scenes. In addition, comparison studies confirm that the proposed TDANet outperforms other state-of-the-art models with higher navigation success rate (SR) and success weighted by length (SPL). TDANet is also deployed in real scenes, showing its superior generalization to the real world.

\section{Related Work}

\subsection{Object-Goal Visual Navigation}
Object-goal visual navigation requires the agent to search for a target instance given only visual observations. Many end-to-end DRL models have been designed to establish the navigation strategy that maps the observations to actions for this task. Some studies learn object-goal visual navigation by learning implicit representations of the observation before inputting it into the navigation policy\cite{baseline, SAVN}. Wortsman \textit{et al.} \cite{SAVN} introduced self-adaptive visual navigation using meta-learning that learns to adapt to test environments without explicit supervision. Other studies exploit the object relationships or semantic contexts aiming for a more robust navigation policy\cite{SP, mjol, VTNet, SpatialAtt, ral, OMT}. For instance, Qiu \textit{et al.}\cite{mjol} and Li \textit{et al.} \cite{Li} integrated hierarchical relationships among objects in the DRL model and achieved remarkable navigation performance using object detection outputs instead of raw RGB images. However, the prior knowledge graph used in these two methods is not updated during training, leading to limited generalizability of the model across different test scenes. Du \textit{et al.} \cite{VTNet} introduced the powerful visual transformer to learn the relationship among objects without the use of a prior knowledge graph. Although improving the navigation performance in unseen test scenes, the visual transformer model significantly complicates computation and training and does not generalize to new classes of unseen objects.

In this letter, TDANet with a novel target attention module is proposed to learn both the semantic and spatial relationships between the target object and observed objects. TDANet is expected to be lightweight and generalizable to both unseen scenes and objects.

\subsection{Zero-Shot Visual Navigation}

Traditional object goal navigation only navigates to limited classes of target objects defined in the training set\cite{SemanticPolicy}. However, new classes of target objects will inevitably appear in the real household environment. Zero-shot navigation, which refers to navigating to objects not selected as the target object in training, has therefore attracted great research interest. While some recent studies, such as CoW\cite{cow} and VoroNav\cite{wu2024voronav}, realized zero-shot navigation with modular designs using large multimodal models, all those methods utilized depth images to generate a global map, which we consider absent from the observations of the agent of interest, thus out of the scope of this letter.

To improve the zero-shot ability of end-to-end DRL navigation models, some efforts have been made through the design of networks \cite{baseline} or input encodings\cite{ZSON, simpleclip, reviewer, SSNet}. For instance, Zhu \textit{et al.}\cite{baseline} proposed a Siamese network\cite{Siamese} for image-goal navigation across different scenes and target images. Khandelwal \textit{et al.}\cite{simpleclip} utilized the zero-shot ability of CLIP\cite{CLIP} to generate semantic embeddings of the goal for navigation. Although the use of CLIP improved zero-shot navigation performance, such a model design has fewer trainable parameters and thus impairs the learning of seen objects in the training set. Xu et al.\cite{reviewer} proposed the Aligning Knowledge Graph with Visual Perception (AKGVP) method leveraging the image-text matching ability of CLIP, achieving remarkable zero-shot navigation capability. Zhao \textit{et al.}\cite{SSNet} proposed SSNet that used cosine semantic similarities and object detection results as the input to the navigation policy to eliminate class-related features. However, the detection matrix used in SSNet is still limited to predefined object classes, impeding further generalization of zero-shot navigation to unseen objects.

In this letter, the Siamese architecture is integrated to TDANet to learn the difference between the desired state of the target object and the state of the observed objects in the current visual observation for zero-shot tasks. Combining the target attention module and Siamese architecture, the proposed TDANet is expected to achieve a strong zero-shot capability with unseen target objects while maintaining satisfactory navigation performance with seen targets.

\section{Task Definition}
In the object-goal visual navigation task, the agent navigates to a target object $t$ defined in the target class set $T$ given only the egocentric RGB image without prior knowledge of the environment. In the zero-shot visual navigation task, the agent learns to navigate to a target object defined in the seen class $S$, and navigates to a target object defined in the unseen class $U$ where $S\cap U=\emptyset$. 

The initial positions of the agent and the target object are randomly selected in each episode. The agent samples its action $a$ through a policy network $\pi$ using the current RGB image $I$ and the word embedding of the target object $w_{t}$ as the input, \textit{i.e.}, ${a\sim\pi_{\theta}\left(I, w_t\right)}$. Here, $\theta$ is the weight of the policy network. $a\in\mathcal{A}= $\{\texttt{MoveAhead}, \texttt{RotateLeft}, \texttt{RotateRight}, \texttt{LookUp}, \texttt{LookDown}, \texttt{Done}\}. The \texttt{MoveAhead} action moves the agent forward 0.25m. The angles of \texttt{Rotate} and \texttt{Look} actions are $45^\circ$ and $30^\circ$, respectively. Finally, the \texttt{Done} action represents the situation when the agent determines that it has found the target and thus ends the episode. An episode is considered a success, when the agent samples the \texttt{Done} action and the target object is \texttt{visible}. The \texttt{visible} indicates that the target object is in the agent's current RGB observation as well as within a distance of 1.5m from it.

\section{Target-Directed Attention Network}

\begin{figure*}[t]
\centerline{\includegraphics[width=1.\linewidth]{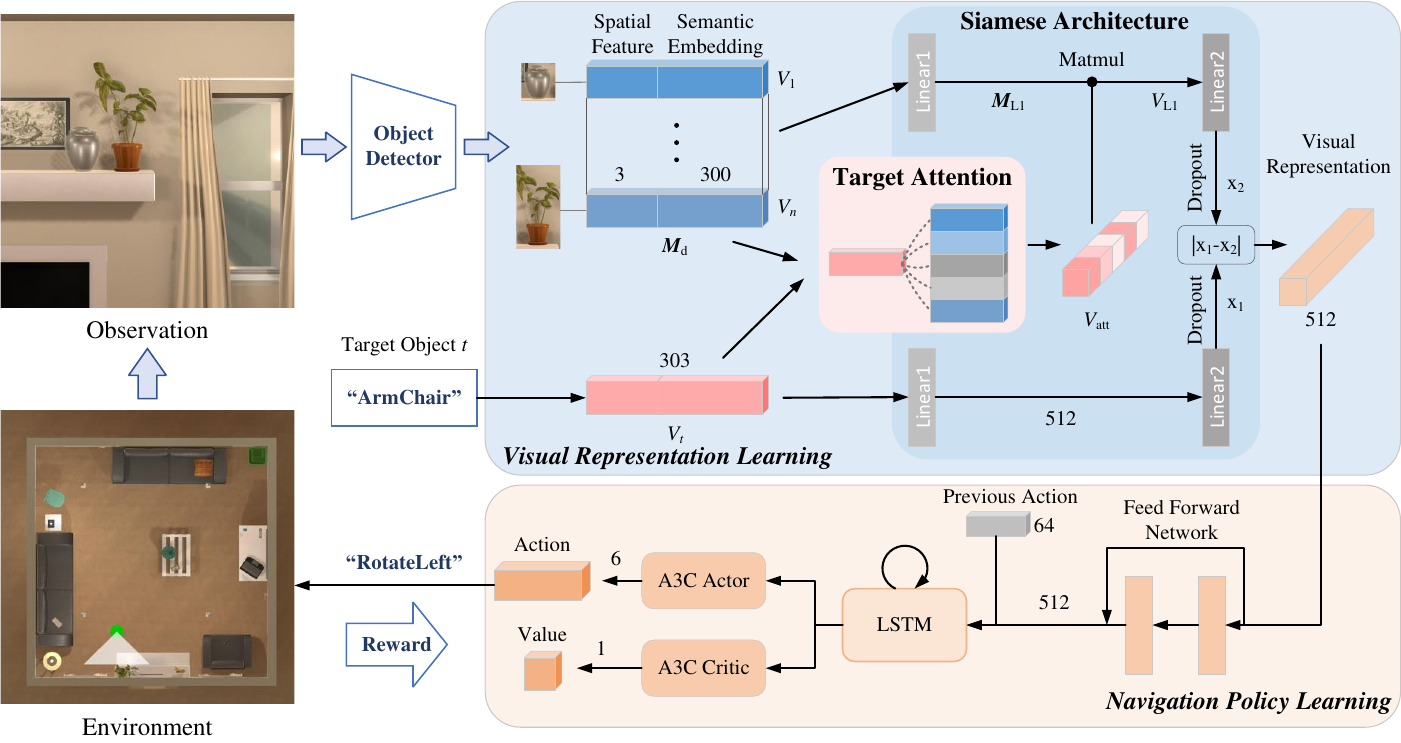}}
\caption{Overview of TDANet. The input is the fused data of bounding boxes and word embeddings of the observed objects and the target object $t$. The target attention module learns the correspondence between the observed objects and the target object $t$ and selects features of the objects most relevant to the target. The Siamese architecture compares the current state with the target state and generates the visual representation. A3C DRL model \cite{a3c} is adopted to learn the navigation policy and is trained with rewards from the environment.}
\label{network}
\end{figure*}

The overview of the proposed target-directed attention network (TDANet) is illustrated in Fig.~\ref{network}. The observation of the agent is the RGB image captured from its monocular camera. The object detector processes the image and generates bounding boxes of objects from the current visual observation. The object detection results with corresponding word embeddings and the word embedding of the target object are inputted into the target attention (TA) module and the Siamese architecture (SA) network to learn the visual representation for the navigation policy. The target attention module learns the spatial and semantic relationships between the observed objects and the target object to select features of objects that have the most correspondence with the target object. The Siamese architecture distinguishes the difference between the observed and target states and encodes it into the visual representation, enabling the zero-shot ability of TDANet. The visual representation is sequentially passed into a feed-forward network and a long short-term memory (LSTM) network to extract deeper features and store the previous memory of navigation. Finally, the asynchronous advantage actor-critic (A3C) model\cite{a3c} learns the navigation policy and controls the agent to navigate to the target object. The network is trained end-to-end and the reward from the environment propagates back to the TA and SA modules, which guides them to learn the meaningful visual representation containing the relationships among objects. The important modules of TDANet are detailed as follows.

\subsection{Input of TDANet}
The input of TDANet consists of two components\textemdash the detected object matrix ${\bm M}_{\rm d}$ and the target vector $V_{\rm t}$. ${\bm M}_{\rm d}$ concatenates the bounding box data of the detected objects with the corresponding word embeddings, \textit{i.e.}, ${\bm M}_{\rm d}=[V_i, i \in \{1, 2, ..., n\}] \in \mathbb{R}^{n \times 303}$, where $V_i=[x_i, y_i, S_i, w_i]$ represents the vector of spatial and semantic information of the detected object $i$. $n$ is the total number of detected objects. $x_i$, $y_i$ and $S_i$ indicate the location $(x, y)$ and the area $S$ of the bounding box of object $i$ in the image coordinates. $w_i$ is the 300-dimensional word embedding of object $i$. $[V_i]$ represents the vertical concatenation of row vectors $V_i$. If there is no object detected in the current frame, ${\bm M}_{\rm d} \in \mathbb{R}^{1 \times 303}$ is set to a zero vector. $V_{\rm t}=[x_{\rm t}, y_{\rm t}, S_{\rm t}, w_{\rm t}]$ is the target vector, representing the desired state of the agent to observe the target object $t$. $x_{\rm t}$, $y_{\rm t}$ and $S_{\rm t}$ are set to 0.5, 0.5 and 0.25, respectively, indicating the desired location of object $t$ is in the middle of the image with an area of a quarter of the entire image area. $w_{\rm t}$ is the word embedding of object $t$. Similar to \cite{mjol, SSNet, OMT, ral}, we use the ground-truth bounding boxes in AI2-THOR and GloVe\cite{glove} to generate word embeddings.

\subsection{Target Attention Module}
The target attention (TA) module is proposed to learn the spatial and semantic correspondence between the detected and target objects. Taking ${\bm M}_{\rm d}$ and $V_{\rm t}$ as inputs, the TA module linearly maps both of them to a vector space of the same dimension and learns the relationships through matrix multiplication, generating correspondence vector $V_{\rm corr}$, \textit{i.e.},

\begin{equation}
V_{\rm corr} = 
\left( V_{\rm t}{\bm W}_{\rm L}+b_{\rm L} \right)
\left( {\bm M}_{\rm d}{\bm W}_{\rm L}+b_{\rm L} \right)^\top
\label{eq_corres}
\end{equation}

\noindent Here, ${\bm W}_{\rm L}\in \mathbb{R}^{d_{\rm input} \times d_{\rm output}}$ and $b_{\rm L}\in \mathbb{R}^{n_{\rm rows} \times d_{\rm output}}$ are the trainable weight and bias of the linear layer, respectively. $d_{\rm input}$ and $d_{\rm output}$ are the input and output dimensions of the linear layer, respectively. $n_{\rm rows}$ is the number of rows of the input matrix. $V_{\rm corr} \in \mathbb{R}^{1 \times n}$ encodes the correspondence of the detected objects with the target object. $n$ is the number of detected objects in the current frame. The TA module then calculates the vector of the attention probability distribution on correspondence values, denoted by $V_{\rm att}$, by applying a softmax function to $V_{\rm corr}$, \textit{i.e.},

\begin{equation}
V_{{\rm att}, i} = \frac{\exp(V_{{\rm corr}, i})}
{\sum_{j}^{n} \exp(V_{{\rm corr}, j})}
\label{eq_att}
\end{equation}

\noindent where $V_{{\rm att}, i}$ and $V_{{\rm corr}, i}$ denote the $i$-th values of $V_{{\rm att}}$ and $V_{{\rm corr}}$, respectively. $n$ is the number of detected objects. 

The vector output of the TA module $V_{\rm att}\in \mathbb{R}^{1 \times n}$ is then multiplied by the extracted feature matrix ${\bm M}_{\rm L1} \in \mathbb{R}^{n \times d_{\rm L1}}$ to obtain ${V}_{\rm L1}$, \textit{i.e.}, ${V}_{\rm L1}=V_{\rm att}{\bm M}_{\rm L1}$. ${\bm M}_{\rm L1}$ represents the extracted features of the $n$ detected objects by passing ${\bm M}_{\rm d}$ through layer ``Linear1" whose output dimension is $d_{\rm L1}$. $V_{\rm L1}$ represents the weighted average feature vector of ${\bm M}_{\rm L1}$ based on the attention distribution. This operation selects the features of detected objects most related to the target object. 

The TA module learns the relationships between the detected objects and the target object during training, expected to help the agent focus on the most relevant objects to the target during navigation, thus improving the navigation success rate and efficiency. In addition, the design of the TA module permits an input of object detection results from an arbitrary number of observed objects. Compared to other work using the one-hot encoding\cite{VTNet} or an input matrix of fixed size \cite{mjol, SSNet}, the proposed TA module allows the update of the detection results when new objects are observed, enhancing the generalization ability.

\subsection{Siamese Architecture}
The Siamese neural network (SNN) is commonly used in few-shot learning tasks such as face recognition\cite{face} and signature verification\cite{signature} where it is difficult to train every instance of the data. SNN usually contains two branches of sub-networks with identical parameters to learn the difference between two inputs. TDANet introduces the Siamese architecture (SA) design to learn the difference between the current and target states, enabling the zero-shot ability of the agent to unseen objects. The SA contains two branches of linear layers sharing the learning weights, as shown in Fig,~\ref{network}. One branch extracts the feature of the detected object matrix ${\bm M}_{\rm d}$ selected by the TA module, and the other extracts the feature of the target vector $V_{\rm t}$. Finally, SA calculates the absolute difference of the output vectors of the two branches with a dropout layer to avoid over-fitting. 

\subsection{Reward Function}
The reward function $R$ is designed as 

\begin{equation}
R = \left \{
\begin{array}{rcl}
R_p  &      &  \text{if }p \text{ is } \texttt{visible}\\
R_t  &      & \text{if }t \text{ is } \texttt{visible} \text{ when } \texttt{Done}\\
R_t+R_p  &      & \text{if }t \text{\&} p\text{ are } \texttt{visible} \text{ when } \texttt{Done} \\
-0.01 &      & \text{otherwise}
\end{array} \right.
\label{eq_reward}
\end{equation}

\noindent Here, $R_t$ and $R_p$ are the target reward and the partial reward, respectively. $t$ and $p$ represent the target object and the parent object, respectively. When the \texttt{Done} action is sampled, the agent receives a target reward $R_{t}=5$ if the target object is \texttt{visible}. We introduce $R_p$ proposed in \cite{mjol} using the parent-target relationship to help the agent learn the relationship among objects. The parent objects are the larger objects related to the target object $t$ in a room. $R_p$ is calculated through $R_{t}\cdot {\rm Pr}\left(t \mid p\right)\cdot k$. Here, $k$ is a scaling factor set to 0.1. ${\rm Pr}\left(t \mid p\right)$ denotes the conditional probability of finding $t$ in the neighborhood of $p$ given the agent observing $p$. It is calculated based on the relative spatial distance of all the parent objects to the target\cite{mjol}. The agent receives $R_p$ when a parent object $p$ is \texttt{visible} in the current RGB frame. In the case of observing the same $p$, the agent does not receive $R_p$ again to encourage exploration. The penalization of $-0.01$ is used to foster a shorter path.

\section{Experiments}

\subsection{Experimental Setup}
We have conducted extensive experiments in the AI2-THOR embodied AI environment to train and test the proposed TDANet. The environment contains 120 near photo-realistic indoor room scenes, including 30 scenes from each of the four room types, \textit{i.e.}, kitchen, bedroom, living room, and bathroom. Following the literature \cite{mjol, SSNet, Li, SAVN}, we select the first 20 rooms from each room type as the training set and the rest 10 rooms from each room type as the test set. The rooms in the test set are unseen during training. The commonly used 22 classes of objects are selected as the target set. 

The agent is trained for 6000,000 episodes on the offline data from AI2-THOR v1.0.1 with a learning rate of 0.0001. During the test, 250 episodes for each room type are evaluated. The evaluation metrics include the success rate (SR) and the success weighted by path length (SPL)\cite{metric}, which are commonly adopted in existing visual navigation studies \cite{SAVN, mjol, SSNet, Li, metric}. SR is calculated as ${\frac{1}{N}\sum^{N}_{i=1}S_i}$, where $N$ is the total number of episodes and $S_i$ is a binary success indicator of the $i$-th episode. SPL is calculated as ${\frac{1}{N}\sum^{N}_{i=1} S_i \frac{L_i}{\max(L_i, e_i)}}$ where $e_i$ is the path length of the agent in the $i$-th episode and $L_i$ is the optimal path length from the agent's initial state to the target object. We evaluate the performance of the trained agents in two sets of episodes where the optimal path length is greater than 1 ($L\geq1$) as well as greater than 5 ($L\geq5$), separately.

\subsection{Comparison Models}

Several benchmark object-goal visual navigation models are selected for comparison, detailed as follows. \textbf{Random}. The agent samples its actions following a uniform probability distribution. \textbf{Baseline}\cite{baseline, SSNet} concatenates the ResNet feature extracted from the current RGB image with the GloVe embedding of the target object as the input. \textbf{SP}\cite{SP} utilizes the prior knowledge of scenes to train the navigation policy. \textbf{SAVN}\cite{SAVN} learns to adapt to new environments during both training and inference using a meta-reinforcement learning approach. \textbf{MJOLNIR}\cite{mjol} introduces a novel context vector in the graph convolutional neural network to learn the hierarchical object relationship. \textbf{Li \textit{et al.}}\cite{Li} combines hierarchical object relationship learning with meta-reinforcement learning, which achieves a state-of-the-art navigation performance in unseen scenes. \textbf{SSNet}\cite{SSNet} is a state-of-the-art model of zero-shot visual navigation, which uses object detection results and cosine similarity of word embeddings as inputs to reduce class-related dependency. 

\subsection{Experiments of Seen Objects in Unseen Scenes}\label{sec_exp}

We train 5 independent agents using TDANet in the training set with all the 22 classes of target objects and deploy them in the unseen scenes in the test set. It takes about 17 hours to train 6M episodes for each agent using three NVIDIA GeForce RTX 4090 GPUs. The average SR and SPL of TDANet in the test set are shown in Table~\ref{tab1} with comparison results from other models using the same experimental setup.

\begin{table}[tbp]
\caption{Comparison results with state-of-the-art Models on seen objects in the test set.}
\renewcommand{\arraystretch}{1.3}
\begin{center}
\tabcolsep=0.037\linewidth
\begin{tabular}{@{}lcccc@{}}
\toprule
\multirow{2}{*}{Model} & \multicolumn{2}{c}{$L\geq1$} & \multicolumn{2}{c}{$L\geq5$}\\
\cmidrule(r){2-3} \cmidrule(r){4-5}
 &  SR (\%) & SPL (\%) &  SR (\%) & SPL (\%)\\
\hline
Random & 11.2 & 5.1 & 1.1 & 0.5 \\
Baseline \cite{baseline} & 35.0 & 10.3 & 25.0 & 10.5 \\
SP \cite{SP} & 35.4 & 10.9 & 23.8 & 10.7 \\
SAVN \cite{SAVN} & 35.7 & 9.3 & 23.9 & 9.4 \\
MJOLNIR-r \cite{mjol} & 54.8 & 19.2 & 41.7 & 18.9 \\
MJOLNIR-o \cite{mjol} & 65.3 & 21.1 & 50.0 & 20.9 \\
SSNet \cite{SSNet} & 63.7 & 22.8 & 42.9 & 21.3  \\
Li \textit{et al.} \cite{Li}& 71.0 & 19.6 & 61.9 & 24.2  \\
\hline
\multirow{2}{*}{\textbf{TDANet (ours)}} & \textbf{78.2} & \textbf{30.6} & \textbf{67.0} & \textbf{33.4} \\
 & $\pm$0.8& $\pm$0.8 & $\pm$1.2&$\pm$0.8 \\
\bottomrule
\end{tabular}
\label{tab1}
\end{center}
\end{table}

TDANet significantly outperforms all the selected models, with an increase of 7.2\% in SR and 7.8\% in SPL when $L\geq1$, and an increase of 5.1\% in SR and 9.2\% in SPL when $L\geq5$. It surpasses the state-of-the-art MJOLNIR and the model proposed by Li \textit{et al.}, both of which require prior construction of a knowledge graph. The results suggest that TDANet successfully learns the spatial and semantic relationships between the target object and the observed objects during training and generalizes well to unseen scenes in the test set. 

\begin{figure}[t]
\centerline{\includegraphics[width=1.0\linewidth]{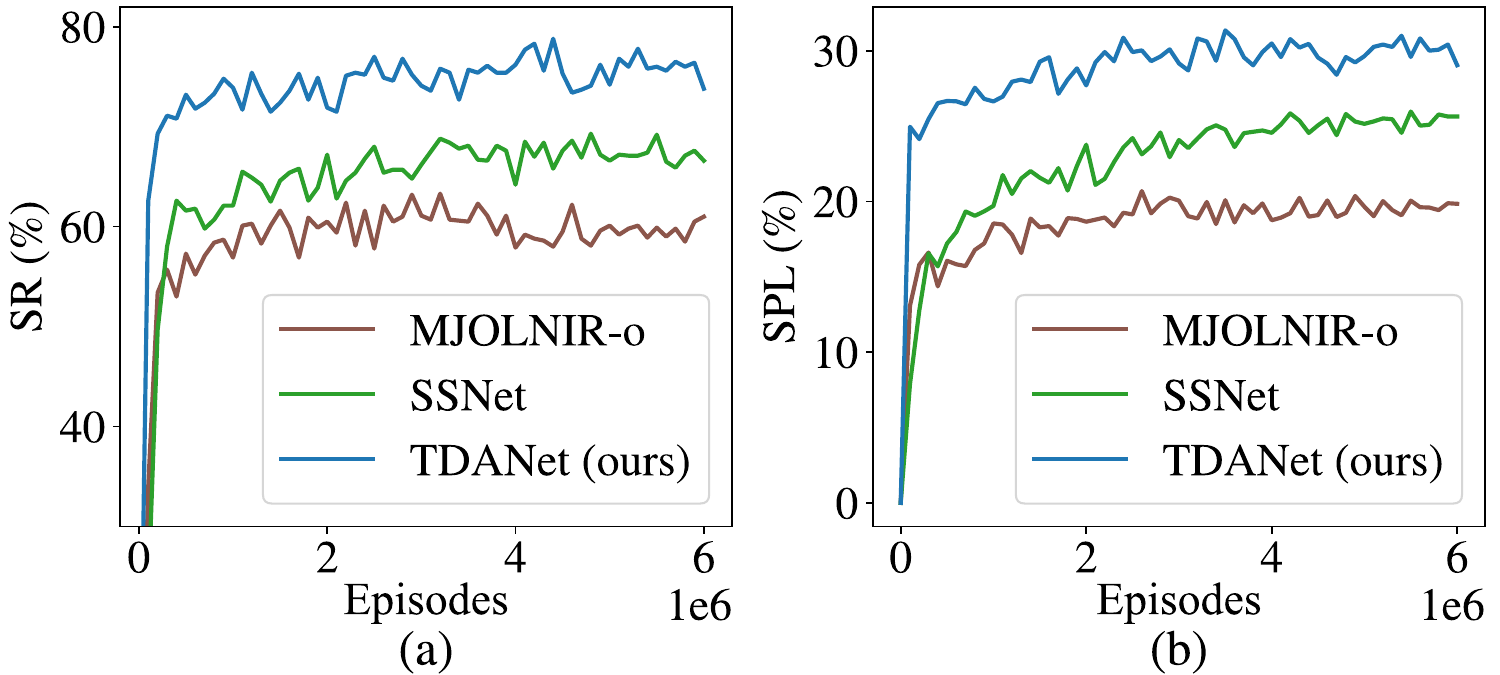}}
\caption{The SR and SPL in the test set evaluated at each training episode of selected comparison models.}
\label{learning_curve}
\end{figure}

The SR and SPL in the test set evaluated at each training episode of different models are shown in Fig.~\ref{learning_curve}. The SR and SPL of TDANet increase more rapidly and reach higher values than other models. Table~\ref{param} lists the numbers of parameters and average inference time of different models calculated using an NVIDIA GeForce RTX 3060 GPU. Although TDANet has more parameters than SSNet, the difference in the inference time between them is only 0.1 ms, which is typically negligible in the task of robot navigation.

\begin{table}[t]
\caption{The numbers of parameters and inference time of different models.}
\renewcommand{\arraystretch}{1.3}
\begin{center}
\tabcolsep=0.08\linewidth
\begin{tabular}{@{}lcc@{}}
\toprule
Model & Param. & Time (ms) \\
\hline
MJOLNIR-o \cite{mjol} & 3.42M & 0.6\\
SSNet \cite{SSNet} & 2.00M& 0.5\\
TDANet(ours) &3.08M & 0.6\\
\bottomrule
\end{tabular}
\label{param}
\end{center}
\end{table}

\begin{figure}[t]
\centerline{\includegraphics[width=1.0\linewidth]{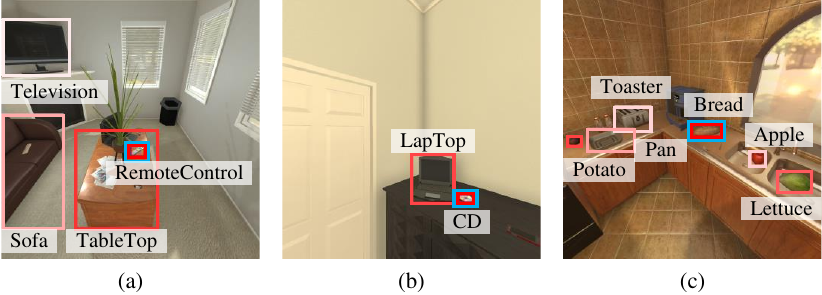}}
\caption{The visualization of the TA module. Only the bounding boxes of objects with higher correspondence values are labeled. The darker red color of a bounding box indicates a higher correspondence value. The target object is marked with the blue bounding box. The results demonstrate that the TA successfully learns the correspondence between the objects and the target.}
\label{relation}
\end{figure}

To analyze the learned relationships using the target attention (TA) module of TDANet, we investigate the correspondence values of observed objects to the target object in the correspondence vector $V_{\rm corr}$ learned using Eq.~(\ref{eq_corres}). Figure~\ref{relation} visualizes the objects of interest with the darker red color representing a higher predicted correspondence value in the current visual observation. We observe that TA consistently predicts the highest value for the target object and higher values for other objects either spatially or semantically more related to the target object. For example, Fig.~\ref{relation}(a) predicts the target object \texttt{RemoteControl} is highly related to the \texttt{TableTop}, \texttt{Sofa} and \texttt{Television}. It suggests that TA pays more attention to the objects where the target is potentially found in their neighborhood, thus improving the navigation success rate and efficiency.

\begin{figure*}[t]
\centerline{\includegraphics[width=0.9\linewidth]{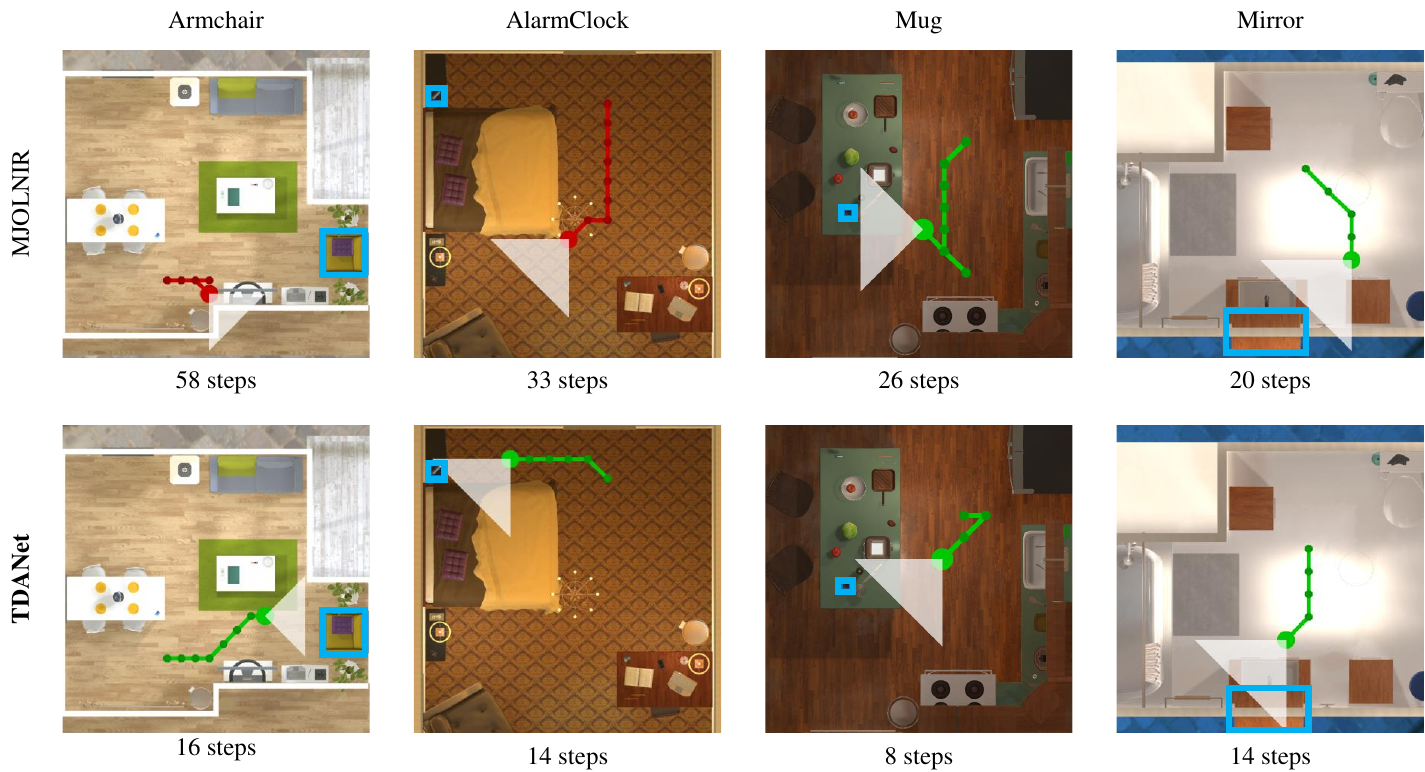}}
\caption{The sampled trajectories of MJOLNIR\cite{mjol} and our TDANet along with the number of navigation steps. Red and green trajectories represent success and failure, respectively. The white triangle indicates the field of view of the agent. The target object is marked with a blue bounding box. }
\label{path}
\end{figure*}

Figure~\ref{path} shows the comparison between the predicted trajectories of MJOLNIR\cite{mjol} and TDANet. TDANet finds the target with the shortest path with a higher success rate, consistent with the experimental results presented in Table~\ref{tab1}.

\subsection{Zero-Shot Experiments}\label{sec_zs_exp}

\begin{table*}[t]
\caption{Comparison results of the zero-shot experiments with state-of-the-art Models in the test set.}
\renewcommand{\arraystretch}{1.3}
\begin{center}
{
\tabcolsep=0.018\linewidth
\begin{tabular}{@{}lccccccccc@{}}
\toprule

\multirow{3}{*}{Model} & 
\multirow{3}{*}{\makecell[c]{Seen/ \\Unseen \\ split}}&
\multicolumn{4}{c}{Unseen class} & 
\multicolumn{4}{c}{Seen class}\\
\cmidrule(r){3-6} \cmidrule(r){7-10}

& & \multicolumn{2}{c}{$L\geq1$} & \multicolumn{2}{c}{$L\geq5$} & \multicolumn{2}{c}{$L\geq1$} & \multicolumn{2}{c}{$L\geq5$}\\
\cmidrule(r){3-4} \cmidrule(r){5-6} \cmidrule(r){7-8} \cmidrule(r){9-10}

& &  SR (\%) & SPL (\%) &  SR (\%) & SPL (\%) &  SR (\%) & SPL (\%) &  SR (\%) & SPL (\%) \\
\hline

Random & \multirow{6}{*}{18/4} & 10.8 & 2.1 & 0.9 & 0.3 & 9.5 & 3.3 & 1.0 & 0.4\\
Baseline \cite{SSNet} &  & 16.9 & 8.7 & 5.3 & 3.1 & 17.7 & 8.3 & 5.3 & 2.6 \\
MJOLNIR \cite{mjol} &  & 20.7 & 7.1 & 10.6 & 4.5 & 51.9 & 16.5 & 33.0 & 14.2 \\
SSNet \cite{SSNet} &  & 28.6 & 9.0 & 12.5 & 5.6 & 59.0 & 19.7 & 38.6 & 18.3 \\
\textbf{TDANet (ours)} & & \textbf{62.5} & \textbf{25.3} & \textbf{47.4} & \textbf{23.8} & \textbf{74.7} & \textbf{29.7} & \textbf{62.9} & \textbf{32.2} \\
\hline

Random & \multirow{6}{*}{14/8} & 8.2 & 3.5 & 0.5 & 0.1 & 8.9 & 3.0 & 0.5 & 0.3 \\
Baseline \cite{SSNet} &  & 14.6 & 4.9 & 4.9 & 2.8 & 30.4 & 9.7 & 11.5 & 5.2 \\
MJOLNIR \cite{mjol} &  & 12.3 & 5.1 & 6.0 & 3.6 & 52.7 & 22.3 & 26.8 & 14.9 \\
SSNet \cite{SSNet} &  & 21.5 & 7.0 & 13.0 & 6.7 & 59.3 & 24.5 & 35.2 & 19.3 \\
\textbf{TDANet (ours)} & & \textbf{53.4} & \textbf{20.5} & \textbf{41.1} & \textbf{20.7} & \textbf{77.1} & \textbf{31.0} & \textbf{64.2} & \textbf{33.6} \\
\bottomrule
\end{tabular}
}
\label{tabzs}
\end{center}
\end{table*}

The 22 target objects are split into seen and unseen object classes similar to \cite{SSNet}. TDANet is first trained to navigate to objects in the seen class in the training set and then deployed in the test set. The detection results of objects in the unseen class are removed and not inputted into the network during training so that the network does not learn any information about the unseen objects. 

Table~\ref{tabzs} shows the evaluation results of the comparison experiments of the zero-shot navigation. It is observed that TDANet achieves the overall best performance with a significant increase of SR and SPL both in the seen and unseen classes. In the task of seen target navigation, TDANet significantly outperforms the state-of-the-art zero-shot model SSNet by an increase of more than 23\% in SR and an increase of 13\% in SPL, when the target object is far from the agent's initial position ($L\geq5$). In the task of unseen target navigation, TDANet surpasses SSNet by a large margin on SR and SPL of more than 28.1\% and 13.5\%, respectively. We conjecture the reason for the improved performance of TDANet over SSNet is as follows. Firstly, The input of SSNet is a fixed-size matrix only containing detection results of a set of predefined objects and cannot update itself with the detection results of unseen objects. In contrast, the TA and SA design of TDANet allows the update of the object detection module with detection results of new objects so that it generalizes significantly better to the navigation task with unseen objects. Secondly, instead of directly using the cosine similarity of word embeddings as SSNet, the TA module adaptively learns the semantic relationships together with spatial relationships to learn domain-independent features. Thirdly, SSNet fuses all features of predefined objects (more than 90 categories), regardless of whether or not it appears in the current observation. In contrast, the TA module only focuses on the most related objects to the target in the current observation, which is more efficient. Fourthly, while SSNet uses cosine similarity as the class-independent data, TDANet predicts actions based on the difference between the current and target states learned by the SA module instead of the certain object class data, resulting in the improved generalization ability.

The results demonstrate that TDANet has robust generalization capabilities in the zero-shot task for unseen objects in unseen scenes while simultaneously maintaining high navigation performance with seen objects during training.

\subsection{Ablation Study}

The ablation study is conducted to evaluate the influence of the target attention (TA) module and the Siamese architecture (SA) of TDANet. For removing the TA module, we calculate the average features of all observed objects. We follow the same experimental setup and evaluation metrics in Section~\ref{sec_exp}. The ablation study results on seen objects are listed in Table~\ref{tab_ablation}. We observe that both the TA and SA modules improve SR and SPL. In addition, TA contributes more to the improvement of navigation performance. We conjecture that TA helps the agent focus more on the spatial and semantic features of the most relevant observed objects to the target, thus improving the success rate and navigation efficiency especially when dealing with trained target objects.

\begin{table}[t]
\caption{Ablation study for TDANet on seen objects in the test set.}
\renewcommand{\arraystretch}{1.3}
\begin{center}
\tabcolsep=0.04\linewidth
\begin{threeparttable}
    \begin{tabular}{@{}cccccc@{}}
    \toprule
    \multicolumn{2}{c}{Module} & \multicolumn{2}{c}{$L\geq1$} & \multicolumn{2}{c}{$L\geq5$}\\
    \cmidrule(r){1-2} \cmidrule(r){3-4} \cmidrule(r){5-6}
    TA \tnote{1} & SA \tnote{2} &  SR (\%) & SPL (\%) &  SR (\%) & SPL (\%)\\
    \hline
    \ding{55} & \ding{55} & 50.0 & 13.9 & 36.8 & 14.7 \\
    \ding{55} & \ding{51} & 58.9 & 15.8 & 41.6 & 15.1 \\
    \ding{51} & \ding{55} & 76.3 & 28.2 & 62.7 & 30.4 \\
    \ding{51} & \ding{51} & 78.8 & 30.6 & 67.7 & 33.4 \\
    \bottomrule
    \end{tabular}
    \begin{tablenotes}
        \footnotesize
        \item[1] TA: Target attention
        \item[2] SA: Siamese architecture 
    \end{tablenotes}
\end{threeparttable}
\label{tab_ablation}
\end{center}
\end{table}

\begin{table}[t]
\caption{Ablation study for TDANet using 18/4 Seen/Unseen class split of zero-shot setting}
\renewcommand{\arraystretch}{1.3}
\begin{center}
\tabcolsep=0.04\linewidth
\begin{threeparttable}
    \begin{tabular}{@{}cccccc@{}}
    \toprule
    \multicolumn{2}{c}{Module} & \multicolumn{2}{c}{Seen class} & \multicolumn{2}{c}{Unseen class}\\
    \cmidrule(r){1-2} \cmidrule(r){3-4} \cmidrule(r){5-6}
    TA \tnote{1} & SA \tnote{2} &  SR (\%) & SPL (\%) &  SR (\%) & SPL (\%)\\
    \hline
    \ding{55} & \ding{55} & 42.1 & 10.6 & 8.1 & 1.2 \\
    \ding{55} & \ding{51} & 56.5 & 15.7 & 17.0 & 4.2 \\
    \ding{51} & \ding{55} & 63.3 & 20.0 & 14.2 & 2.1 \\
    \ding{51} & \ding{51} & 74.7 & 29.7 & 62.5 & 25.3 \\
    \bottomrule
    \end{tabular}
    \begin{tablenotes}
        \footnotesize
        \item[1] TA: Target attention
        \item[2] SA: Siamese architecture 
    \end{tablenotes}
\end{threeparttable}
\label{tab_ablation_zero}
\end{center}
\end{table}

\begin{figure}[t]
\centerline{\includegraphics[width=0.85\linewidth]{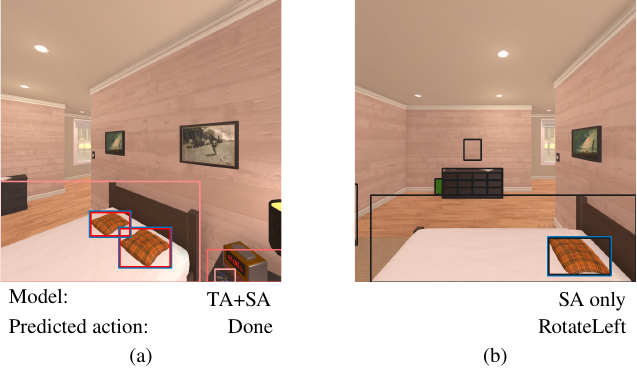}}
\caption{The comparison of TDANet and the SA-only network for unseen object goal navigation. The target object \texttt{Pillow} is marked with the blue bounding box. (a) TDANet predicts the right action by focusing on objects related to the target. (b) Without the help of the TA module, the SA-only network is distracted by unrelated objects and predicts the wrong action.}
\label{fig_ablation}
\end{figure}

To analyze the zero-shot performance of TDANet, another ablation study under the same experimental setup in Section~\ref{sec_zs_exp} is conducted and the results of the zero-shot setting are listed in Table~\ref{sec_zs_exp}. It is observed that the combination of the TA and SA modules significantly increases SR and SPL by more than 40\% and 20\% in the zero-shot task, respectively. The design of the TA and SA modules significantly improves the zero-shot ability of TDANet. The TA module selects features of the observed objects related to the target, which are inputted into the SA module to learn the difference between the target and current states. TDANet then predicts actions based on the class-independent features outputted from the SA module, thus improving the zero-shot navigation ability. In comparison, When using the SA module only, the average feature of all observed objects is calculated and inputted into the SA module. As illustrated in Fig.~\ref{fig_ablation}, the SA-only network is distracted by unrelated objects without the help of the TA module and predicts wrong actions when it finds the target, while TDANet predicts the right action by focusing on the objects related to the target.

\subsection{Real-world Deployment}

\begin{figure}[t]
\centerline{\includegraphics[width=1.0\linewidth]{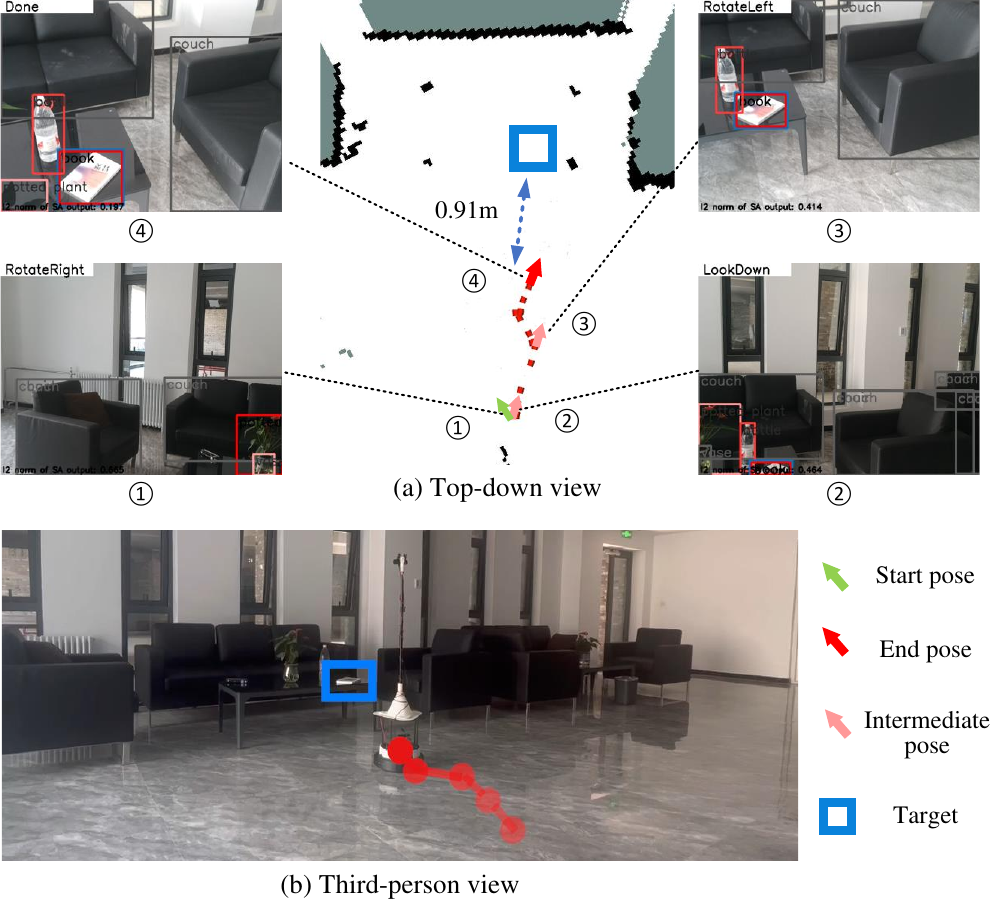}}
\caption{A sample trajectory of the real-world deployment of TDANet to the unseen object \texttt{Book}. (a) Top-down view of the trajectory. {\ding{172}}-{\ding{175}} are sampled egocentric RGB observations of the robot marked with detected bounding boxes of YOLOv7 and the predicted action of TDANet at different poses of the trajectory. The darker red color of the bounding box represents a higher correspondence value predicted by the TA module. (b) Third-person view of the scene and the robot trajectory.}
\label{fig_real_world}
\end{figure}

A testing system using a TurtleBot4 wheeled robot is developed and an OAK-D-Pro camera is installed on the robot at a height of 1.5m above the ground to test the real-world generalization of TDANet. A servo is equipped to control the camera's rotation. Navigation only required the RGB frame of the camera. The agent trained in Section~\ref{sec_exp} is used to deploy and YOLOv7\cite{yolov7} pretrained in the COCO dataset is used as the object detector of TDANet. All algorithms run in real-time on a laptop with an i7-12700H CPU and an NVIDIA GeForce RTX 3060 GPU. A sample navigation trajectory of the agent to an unseen object is visualized in Fig.~\ref{fig_real_world}. Videos are available in the supplementary items. TDANet successfully navigates the robot to the unseen target object \texttt{Book} by focusing on the observed objects that are most related to the target in the current observation. For example, in Fig.~\ref{fig_real_world}\ding{172}, \texttt{PottedPlant} in the bottom right corner of the image is predicted as the most related object and an action of \texttt{RotateRight} is predicted. In Fig.~\ref{fig_real_world}\ding{173}, TDANet locates the target at the bottom of the image and predicts the action \texttt{LookDown}. The experimental results illustrate the satisfactory generalization capability of TDANet to the real world.

\section{Conclusion}

This letter proposed the target-directed attention network (TDANet), which paid more attention to the most relevant objects to the target object in the monocular visual observation during navigation. A target attention module was designed to learn the spatial and semantic correspondence of the observed objects with the target object. The adopted Siamese architecture compared the current state to the target state, improving the generalization of TDANet. Extensive comparison experiments and ablation studies in the AI2-THOR environment were conducted, the results of which demonstrated that TDANet learned a domain-independent visual representation for navigation policy with a strong generalization ability to both unseen scenes and unseen target objects, achieving higher navigation success rate and efficiency compared to other selected state-of-the-art models. The deployment of TDANet in the real world demonstrated its generalization ability in real-world environments.

In future work, we plan to investigate the collision avoidance of TDANet in more complex simulated environments\cite{doze} for a safe application in the real world.

\balance

\end{document}